\documentclass[letterpaper, 10 pt, conference]{ieeeconf}  

\IEEEoverridecommandlockouts                              

\usepackage{cite}
\usepackage{amsmath,amssymb,amsfonts}
\usepackage[dvipsnames]{xcolor}
\usepackage{hyperref}
\usepackage{booktabs}
\usepackage{makecell}

\usepackage{amsthm}
\usepackage[binary-units]{siunitx}[=v2]
\usepackage{mathtools}
\usepackage{tikz}
\usepackage{pdfpages}
\usepackage[T1]{fontenc}
\usepackage{bm}
\usepackage{forest}
\usepackage{xfrac}
\usepackage{amsmath,amssymb,graphicx}
\usepackage{tabularx}
\usepackage{multirow}
\usepackage{adjustbox}
\usepackage{floatrow}
\usepackage{caption}

\DeclareRobustCommand\mytikzLineRed{\tikz[baseline=-0.75ex]\draw[line width=2pt,color=red] (0,0) -- (0.2,0);}
\DeclareRobustCommand\mytikzLineGreen{\tikz[baseline=-0.75ex]\draw[line width=2pt,color=ForestGreen] (0,0) -- (0.2,0);}
\DeclareRobustCommand\mytikzLineOrange{\tikz[baseline=-0.75ex]\draw[line width=2pt,color=orange] (0,0) -- (0.2,0);}
\DeclareRobustCommand\mytikzLineBlue{\tikz[baseline=-0.75ex]\draw[line width=2pt,color=blue] (0,0) -- (0.2,0);}
\DeclareRobustCommand\mytikzLineBlack{\tikz[baseline=-0.75ex]\draw[line width=2pt,color=black] (0,0) -- (0.2,0);}

\definecolor{BlueRawignal}{HTML}{0000a2}

\definecolor{RedRawignal}{HTML}{F28522}

\definecolor{MagentaReducedSignal}{HTML}{800074}

\definecolor{GreenReducedSignal}{HTML}{06592A}

\def\BibTeX{{\rm B\kern-.05em{\sc i\kern-.025em b}\kern-.08em
    T\kern-.1667em\lower.7ex\hbox{E}\kern-.125emX}}

\title{\LARGE \bf
A Data-Driven Novelty Score for Diverse In-Vehicle Data Recording }

\author{Philipp Reis$^{1}$, Joshua Ransiek$^{1}$, David Petri$^{2}$, Jacob Langner$^{1}$ and Eric Sax$^{2}$
\thanks{$^{1}$Philipp Reis, Joshua Ransiek and Jacob Langner with the FZI Research Center for Information Technology, 76131 Karlsruhe, Germany
        {\tt\small \{reis,ransiek,langner,sax\}@fzi.de}, Philipp Reis is corresponding author}%
\thanks{$^{2}$ David Petri and Eric Sax are with the Karlsruhe Institute of Technology, 76131, Karlsruhe, Germany}
}

\begin{document}
\theoremstyle{definition}
\newtheorem{definition}{Definition}
\renewcommand{\theadalign}{vh}
\newcommand{\probP}{\text{I\kern-0.15em P}}

\newcommand{\mathdefault}[1][]{}

\maketitle
\thispagestyle{empty}
\pagestyle{empty}

\begin{abstract}
High-quality datasets are essential for training robust perception systems in autonomous driving. However, real-world data collection is often biased toward common scenes and objects, leaving novel cases underrepresented. This imbalance hinders model generalization and compromises safety. The core issue is the curse of rarity. Over time, novel events occur infrequently, and standard logging methods fail to capture them effectively. As a result, large volumes of redundant data are stored, while critical novel cases are diluted, leading to biased datasets.
This work presents a real-time data selection method focused on object-level novelty detection to build more balanced and diverse datasets. The method assigns a data-driven novelty score to image frames using a novel dynamic Mean Shift algorithm. It models normal content based on mean and covariance statistics to identify frames with novel objects, discarding those with redundant elements.
The main findings show that reducing the training dataset size with this method can improve model performance, whereas higher redundancy tends to degrade it. Moreover, as data redundancy increases, more aggressive filtering becomes both possible and beneficial. While random sampling can offer some gains, it often leads to overfitting and unpredictability in outcomes.
The proposed method supports real-time deployment with 32 frames per second and is constant over time. By continuously updating the definition of normal content, it enables efficient detection of novelties in a continuous data stream.

\end{abstract}


\section{Introduction}
\label{sec:introduction}

The development of highly automated driving (HAD) systems depends on the continuous acquisition of large-scale data to train and validate machine learning models. A core challenge in this process is the volume-value trade-off. While modern test vehicles produce up to $\SI{20}{\giga\bit\per\second}$ of raw data during operation~\cite{heinrich24}, much of it is redundant and offers little value in improving system performance~\cite{Liu2024}. To comply with safety standards like ISO 8800~\cite{ISO8800}, data selection must prioritize quality and ensure a balanced dataset of novel and normal elements to avoid bias \cite{Langner23,10005441}.
Prioritizing novel events during data collection is critical for enhancing system robustness~\cite{Breitenstein20}. However, novelty detection faces a foundational issue: the definition of what is \textit{normal}. Existing benchmarks often define all Cityscape classes as \textit{normal}~\cite{Blum2021, Chan21, Hendrycks2022}, which fails to capture the open-world variability inherent in real-world driving environments. Predefined notions of normality are static and incomplete. To address this, a data-driven approach to defining normal and novel instances has been proposed~\cite{bogdoll24}.
This paper introduces a method for real-time, object-level novelty detection based on a Mean Shift algorithm. The approach dynamically models normality using incoming data and identifies deviations as novel. By continuously updating this model, the system adapts to environmental changes and directs data collection toward novel objects. The goal is to reduce redundancy, capture more informative samples, and support the development of safe AI systems.

\begin{figure}[t]
    \centering
    \includegraphics[width=0.7\linewidth]{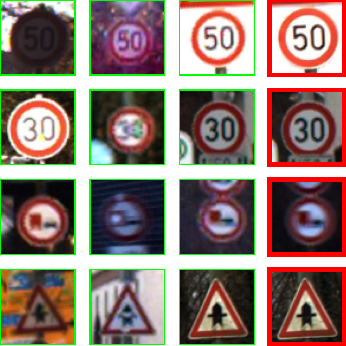}
    \caption{Examples of different road sign classes, with each row showing three instances followed by a redundant example (red frame) that is not recorded due to similarity with the preceding samples.}
    \label{fig:IntroEye}
\end{figure}

\subsection{Related Works}

Novelty detection in streaming image data is critical for real-time applications such as HAD. Existing methods can be grouped into probabilistic modeling, distance-based approaches, reconstruction-based techniques, and information-theoretic strategies, each with notable strengths and limitations.
Probabilistic methods, such as Bayesian neural networks and normalizing flows, estimate sample likelihoods~\cite{Kendall2015,grcic2021}. Also, Dirichlet Prior Networks \cite{malinin2018} model predictive uncertainty over class distributions. However, these methods are computationally expensive or unsuitable for real-time adaptation. Distance-based approaches are presented in~\cite{Zimek2015,Knorr98,Perera2019}, which measure deviations from known data. Reconstruction-based techniques, leveraging autoencoders or GANs \cite{Schlegl2017,Hofmockel2018, SCHLEGL2019}, detect anomalies via reconstruction errors, yet struggle with distributional shifts and often demand costly retraining. Information-theoretic strategies like Contrastive Shifted Instances \cite{Tack2020} and Outlier Exposure \cite{Hendrycks2022} offer contrastive learning-based novelty detection but depend heavily on large-scale pretraining and are not inherently adaptive.
Furthermore, dataset reduction using optimization techniques has been proposed in~\cite{Wei2015} and applied in~\cite{Kaushal2018}, but these methods are generally less effective than uncertainty-based approaches, as shown in~\cite{Birodkar2019}. Gradient-based methods~\cite{lee2017,Guo2022DeepCoreAC,Mirzasoleiman20,killamsetty2021} aim to approximate the gradient of a (sub)set by selecting data points that closely match the overall gradient direction. However, all of these approaches assume access to a static dataset, making them unsuitable for dynamic data streams where the full dataset is not available at once.

A common shortcoming across these methods is the lack of lightweight, continuously adaptive frameworks suitable for evolving data streams. Most approaches either retrain extensively, require high computational resources, or maintain complex external memories.

\subsection{Contribution and Outline}
As a complement to earlier work that operates on time series data \cite{reis2024},  this work introduces a lightweight, dynamic streaming novelty detection method on images based on a mean shift model. Unlike existing approaches that rely on static thresholds or computationally expensive model retraining, the proposed method operates directly in semantic embedding space derived from a pretrained EfficientNet model. Incoming image samples are embedded and compared to a continuously updated empirical mean and covariance. Samples with Mahalanobis distances exceeding a defined threshold are flagged as novel and incorporated into the distribution model, enabling dynamic refinement of what is considered \textit{normal}.
The method supports real-time execution at up to 32 frames per second (FPS), even under worst-case conditions, with minimal computational overhead. It enables efficient novelty detection without large memory buffers or frequent inference passes, making it well-suited for resource-constrained, high-throughput applications such as HAD and online learning systems.
The key contribution are:
\begin{itemize}
    \item \textbf{Data-Driven Novelty Scoring:} A novel Mean Shift algorithm assigns dynamic novelty scores based on previously seen samples, using continuously updated mean and covariance statistics. 
    \item \textbf{Redundancy-Aware Filtering:} Filtering out redundant data improves generalization and reduces computation; redundancy correlates with degraded model performance.
    \item \textbf{Sampling Strategy Analysis:} Novelty filtering outperforms random sampling, which often causes overfitting and inconsistent results.
    \item \textbf{Dataset Balance and Diversity Evaluation}: Novelty filtering improves dataset balance and diversity. Too aggressive filtering can lead to overfitting in downstream training tasks. 
\end{itemize}

Section~\ref{sec:preleminaries} outlines the problem for dynamic object level novelty detection, Section~\ref{sec:MeanShift} describes the mean-shift algorithm to detect novel objects in an image. The proposed dynamic approach is presented in Section~\ref{sec:AdaptiveMeanShift} with quantitative evaluations regarding model performance and the resulting data balance and diversity. in Section~\ref{sec:evaluation}. Section~\ref{sec:conclusion_and_outlook} concludes the results of this contribution.

\section{Problem Setup}\label{sec:preleminaries}

The general problem is to detect novel objects $\bm{o}_\mathrm{nov}$ in a video stream $\mathcal{D}$.
A video steam consists of incoming frames $\bm{f}\in \mathcal{D}$, where each frame is of height $\mathrm{h}$, width $\mathrm{w}$ and number of channels $\mathrm{c}$, so that $\bm{f}\in\mathbb{R}^{\mathrm{w}\times \mathrm{h}\times \mathrm{c}}$.
Novelty detection on the object level involves identifying novel objects $\bm{o}_\mathrm{nov}$ within a frame that deviates from the distribution of normal objects $\bm{o}_\mathrm{norm}$ which have been already collected.  It is assumed that objects are within a sub-frame, called patch $\bm{p}$, and is defined by its upper left position  $(x_i,y_i)$ and its patch sizes $(w_\mathrm{p},h_\mathrm{p})$ within the frame, so it follows
\begin{equation*}
   \mathcal{P}  = \{\bm{p}_i\}_{i=1}^\mathrm{S}, 
   \quad\bm{p}_i=(x_i,y_i,w,h),
\end{equation*}
where $\mathrm{S}$ is the number of objects in a frame resulting in $\mathrm{S}$ patches.
Each extracted patch $\bm{p}_i$ is mapped to a feature vector $\bm{z}_i\in\mathcal{Z}$  through a feature extraction function $\bm{\phi}$, so that
\begin{equation*}
    \bm{z}_i = \bm{\phi}(\bm{p}_i),\quad \bm{z}_i\in \mathbb{R}^\mathrm{d}.
\end{equation*}
For detecting novelties, the characteristics of \textit{normal} objects $\bm{o}_\mathrm{norm}$ in previously observed frame patches are considered. This characterization evolves over time as more data is encountered, enabling the detection process to adapt dynamically, which means
\begin{subequations}
\begin{equation*}
    \bm{o}_\mathrm{norm}(t) = \mathcal{F}_\mathrm{norm}(\bm{p}_i(t),t)
\end{equation*}
\begin{equation*}
    \bm{o}_\mathrm{nov}(t) = \mathcal{F}_\mathrm{\mathrm{nov}}(\bm{p}_i(t),t).
\end{equation*}
\end{subequations}
Consequently, the definition of a \textit{novel} object~$\bm{o}_\mathrm{nov}$ is data-driven, relying on a continuous learning process to refine the distinction between \textit{normal} objects~$\bm{o}_\mathrm{norm}$ and \textit{novel} objects~$\bm{o}_\mathrm{nov}$.
Within this contribution, the goal is to detect \textit{normal} objects~$\bm{o}_\mathrm{norm}$ over time, and learning their representation during execution to dynamically adapt the definition of \textit{normal} objects~$\bm{o}_\mathrm{norm}$. This is an essential task for dynamic environments as it appears in automotive perception.
\section{Static Mean-Shift Algorithm}\label{sec:MeanShift}
\begin{figure*}[t]
    \centering
        \includegraphics[width=0.95\linewidth]{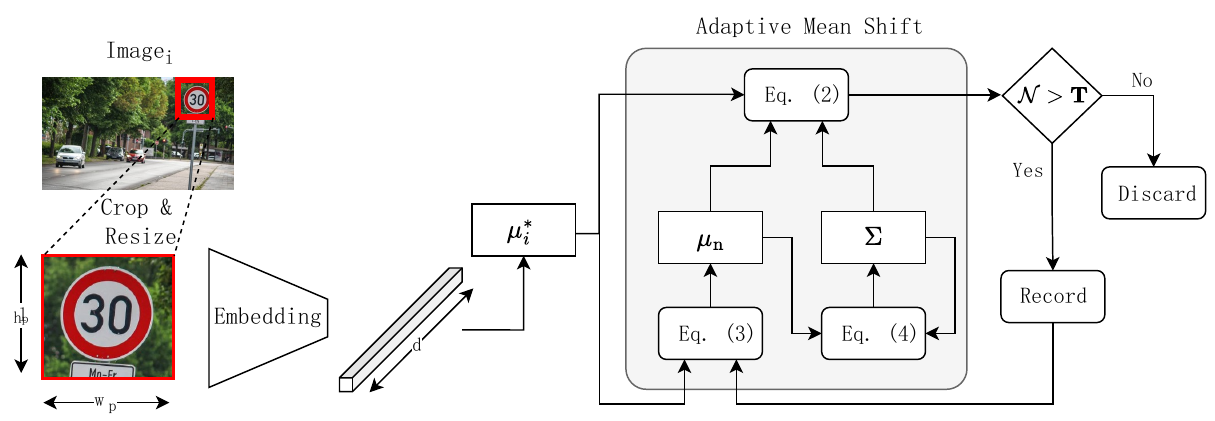}
        \caption{Concept of the dynamic Mean Shifts for Novelty Detection in image frames. First, object of interests are cropped and resized. These patches are embedded and evaluated in the dynamic mean shift algorithm for novelties. In case of a Novelty Score $\mathcal{N}>\mathbf{T}$ the image is considered as a novelty and the image will be recorded and the patch is updated into the statistics as normal. Otherwise, the image will be discarded. }
        \label{fig:ConceptMeanShift}
\end{figure*}
For the detection of novelties, the Mean-Shift Algorithm is used  which is based on a static training data set as proposed in \cite{hermann22}. This approach uses the Hotelling $T^2$ test, which estimates the significance of mean-shifts between two populations. This contribution labels a frame tested for novelty with an asterisk $\bm{}^\ast$. For the definition of normal, objects of interest are collected as single patches. These patches are then resized to a predefined patch size $(w_\mathrm{p},h_\mathrm{p})$. Those patches are then mapped onto a flattened intermediate feature vector $\bm{z}_{i} \in \mathbb{R}^\mathrm{d}$ by a neural network $\Phi$. The size $\mathrm{d}$ depends on the size of the chosen intermediate layer.
From flattened feature representation, the mean is computed as 
\begin{equation*}
    \label{eq:mu_N}
     \bm{\mu}_{\mathrm{n}} = \frac{1}{N} \sum_{i=1}^{N} z_{i}.
\end{equation*}
This represents the mean value of the \textit{normal} data set.  
The covariance of the normal data $\bm{X}$ is given by 
\begin{equation*}
    \label{eq:cov_mean_shifts}
    \bm{\hat{\Sigma}} = \frac{1}{N-1} (\bm{X} - \bm{\mu}_{\mathrm{n}})^\intercal (\bm{X} - \bm{\mu}_{\mathrm{n}}),
\end{equation*}
which is the empirical covariance matrix of the training dataset $\mathcal{D}$. For robustness reason, the covariance estimator is computed using the Ledoit–Wolf shrinkage concept \cite{ledoit04}.  The estimator is a  convex combination of a scaled identity matrix $\mathbb{I}$ and the empirical covariance matrix weighted with the  shrinkage factor $\alpha\in [0,1]$ 
\begin{equation*}
    \bm{\Sigma} = (1-\alpha)\hat{\Sigma}+\alpha\frac{\mathrm{tr}(\hat{\Sigma})}{D}\mathbb{I}.
\end{equation*}
The shrinkage factor $\alpha$ is given analytically by minimizing the quadratic loss between the true and estimated covariance matrix \cite{ledoit04}. 

For detecting novelties the flattened feature representation $z_{i}^\ast =\bm{\phi}(\bm{p}^\ast{i})$ of a test frame $\bm{f}^\ast$ is centered $\mu_\mathrm{c}^\ast = \mu^\ast-\mu_\mathrm{n}$ and given the 
covariance $\bm{\Sigma}$ of the normal data set, the novelty score $\mathcal{N}$ is defined by the mean shifts using the unnormalized Hotelling $\tilde{T}^2$ test statistics, given by
\begin{equation}
    \label{eq:hotelling_squared}
    \mathcal{N}(\bm{\mu}_i^\ast,\bm{\mu}_n,\bm{\Sigma}_n) = \tilde{T}^2= (\bm{\mu^\ast}_i - \bm{\mu}_{\mathrm{n}})^\top \bm{\Sigma}^{-1} (\bm{\mu^\ast}_i - \bm{\mu}_{\mathrm{n}}).
\end{equation}
This is visually the Mahalanobis distance of a sample to the dataset distribution.
In case the novelty score $\mathcal{N}$ is greater than a predefined threshold $\mathcal{N}>\mathbf{T}$, this given patch is considered as a novelty.

\section{Concept of the Dynamic Mean-Shift for Novelty Detection} \label{sec:AdaptiveMeanShift}
The nature of a dynamic environment in vehicle perception necessitates a dynamic approach to detecting novelties. A concept for the dynamic novelty detection can be seen in Fig.  \ref{fig:ConceptMeanShift}.
The goal is to only record frames of a video stream that include novel objects compared to already captured data. The images in a data stream are evaluated for objects of interest as image patches, embedded one by one, and compared to the normal data set using the mean-shift algorithm. The output is a novelty score $\mathcal{N}$, which decides to record or discard the image. If the novelty score $\mathcal{N}$ is below the threshold $\mathbf{T}$, the image is considered redundant and is discarded and the next image can be evaluated. In case of a novelty, the normal data set will be updated with this patch. This updated normal data set is the basis for further comparison in the mean shift algorithm, representing the dynamic novelty detection. 

Since the mean shift algorithm described in section \ref{sec:MeanShift} works on a static definition of a \textit{dataset}, a dynamic representation of a normal dataset in the form of an dynamic mean $\bm{\mu}_{n}(t)$ and $\bm{{\Sigma}_{n} }(t)$ is necessary. To utilize this approach as a method to build a continuously growing normal dataset, those values need to be calculated with every new image that contains a novelty and is, therefore, updated into the normal dataset.  To make those calculations more efficient, a reformulation to update after $K$ novelties is possible but not necessary in most cases, as discussed in section \ref{sec:runtime}. To be more general, it is assumed $K$ novelties are updated, but for single updates $K=1$. By only taking the current value of the feature vector $\bm{\mu}_{\mathrm{n}}$, the number $N$ of already recorded normal instances and the  feature vector $\bm{\mu}^\ast$ of the $K$ new instances as input the mean feature vector $\bm{\mu}_{\mathrm{n}}$ can be updated without access to all individual feature vectors of all already recorded patches by
\begin{equation}
    \bm{\mu}_\mathrm{n} = \frac{1}{(N+K)} \bigg[N \bm{\mu}_{\mathrm{n}} + K \bm{\mu}_\mathrm{K}\bigg].
    \label{eq:cont_mu_short}
\end{equation}
Similarly, for the covariance matrix \eqref{eq:cov_mean_shifts} with  $\mathrm{M}=\mathrm{N}+\mathrm{K}$ as the total number of novel and normal patch samples follows
\begin{multline}
    \bm{\Sigma}_\mathrm{} = \frac{1}{\mathrm{M}-1} \bigg[\bm{A_\mathrm{M}}
    - \bm{\mu}_\mathrm{M}\bm{b_\mathrm{M}}^\intercal \\
    - \bm{b}_\mathrm{M}\bm{\mu}_\mathrm{M}^\intercal 
    + \mathrm{M}\cdot\bm{\mu}_\mathrm{M}\bm{\mu}_\mathrm{M}^\intercal \bigg].
    \label{eq:cont_cov}
\end{multline}
The sums $\bm{A}_\mathrm{M}$ and $\bm{b}_\mathrm{M}$ are calculated by
\begin{equation*}
\begin{split}
    \bm{A}_\mathrm{M} &= \bm{A}_\mathrm{N} + \sum_{i=N+1}^{\mathrm{M}}\bm{z_i}\bm{z_i}^\intercal\\
    \bm{b_\mathrm{M}} &= \bm{b_\mathrm{N}}+\sum_{i=N+1}^\mathrm{M}  \bm{z_i}. 
\end {split}
\end{equation*}
The proof for the dynamic update of the mean and the covariance is given in Appendix A1 and A2.
After this update, the next image can be evaluated for objects of interest.

\section{Numerical Evaluation}
\label{sec:evaluation}
For the evaluation of the dynamic mean shift algorithm, the German Traffic Sign Benchmark (GTSB) dataset \cite{Stallkamp2011} is used consisting of 8164 images for training and 12629 for testing across 42 traffic sign classes.

\subsection{Run Time Analysis}\label{sec:runtime}

\begin{table}[t]
    \centering
    \caption{Run time analysis for the novelty detection algorithm for different patch sizes and number of patches averaged over 8165 computations and rounded to the fourth decimal. The analysis is computed using a NVIDIA GTX 2080 graphics processing unit and an Intel Core i9-9900K CPU with a clock speed of 3.60 GHz.}
    \label{tab:run_time_table}
    
\resizebox{\columnwidth}{!}{%
\begin{tabular}{lcccccc}
\toprule\toprule
\multirow{2}{*}{} & \multirow{2}{*}{\# Patches} & \multicolumn{3}{c}{Run times [ms] for different Patch Sizes} \\
\cmidrule(lr){3-5}
 &  & $64^2$ & $128^2$ & $256^2$ \\
\midrule
\multirow{4}{*}{\shortstack{Novelty\\Score}} 
    & 1  & $12.85\pm0.1$    & $12.88\pm0.1$     & $13.02\pm0.2$ \\
    & 4  & $13.15\pm0.3$    & $13.20\pm0.2$     & $13.39\pm0.3$ \\
    & 16 & $13.39\pm0.3$    & $14.01\pm0.3$     & $14.41\pm0.3$ \\
    & 64 & $14.33\pm0.2$    & $14.69\pm0.2$     & $14.83\pm0.2$ \\
\midrule
\multirow{4}{*}{\shortstack{Covariance\\Update}}
    & 1 & $15.77\pm0.1$ & $76\pm0.4$ & $8660\pm21$ \\
    & 4 & $17.05\pm0.3$ & $77\pm1.1$ & $8420\pm17$ \\
    & 16 & $18.61\pm0.4$ & $79\pm1.3$ & $8701\pm23$ \\
    & 64 & $18.33\pm0.4$ & $79\pm2.2$ & $8880\pm28$ \\
\bottomrule
\end{tabular}
}

\end{table}
To investigate the computational efficiency of the dynamic mean shift novelty detection algorithm, an experiment to analyze the runtime is conducted. For the run time analysis, the frame was rescaled to different frame sizes and different numbers of patches per batch, see Tab.~\ref{tab:run_time_table}. Three main results can be derived. First, the novelty score computation is weakly insensitive regarding the patch size and number of patches. In worst-case scenarios with 64 objects of interest with a patch size of $256^2$, 71 FPS is possible. The second result is that the update can be efficient but is strongly dependent on the patch size, while the impact on the number of patches is moderate. As a result, for high-frequency updates in the area of milliseconds, smaller patch sizes are necessary. For real-time requirements, a patch size of $64^2$ is recommended. In this case, a 32 FPS can be scored and updated as a worst-case scenario with 64 novelties per image.
The third main result is the stable computation time. Since the definition of normal is based on a mean and covariance statistic, the novelty score computation, as well as the statistics update, stay stable over time. This is an advantage over distance- or density-based novelty detection methods, as the calculation time increases with the set size of normal images or must be shrunk regularly.

\subsection{Dynamic Novelty Detection on Automotive Data streams}
\subsubsection{Implementation Details}
The proposed dynamic novelty detection method processes the incoming patches and resizes them to a size of $64^2$. The patches are embedded using the EfficientNet-B4 network and hooked after the fifth block, resulting in a feature matrix of $160\times4\times4$, which is flattened to a feature vector of length $z_i=2560$. The mean $\mu_\mathrm{n}$ and the covariance $\Sigma$ are initialized with a gaussian noise as \textit{normal} data set. 
\subsubsection{Data Set Reduction using the Novelty Detection}\label{sec:dataReduction}
\begin{figure}[t]
    \centering
    \includegraphics[width=0.90\linewidth]{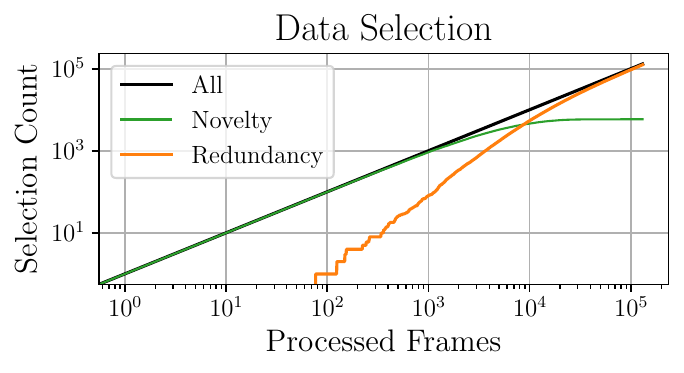}
    \caption{Exemplaric data selection over frame index in double logarithmic scale of all selected data (\mytikzLineBlack), the selected novelty elements (\mytikzLineGreen), and resulting redundant elements (\mytikzLineOrange).}
    \label{fig:selection_process}
\end{figure}
\begin{figure*}[t]
    \centering
    \includegraphics[width=0.95\linewidth]{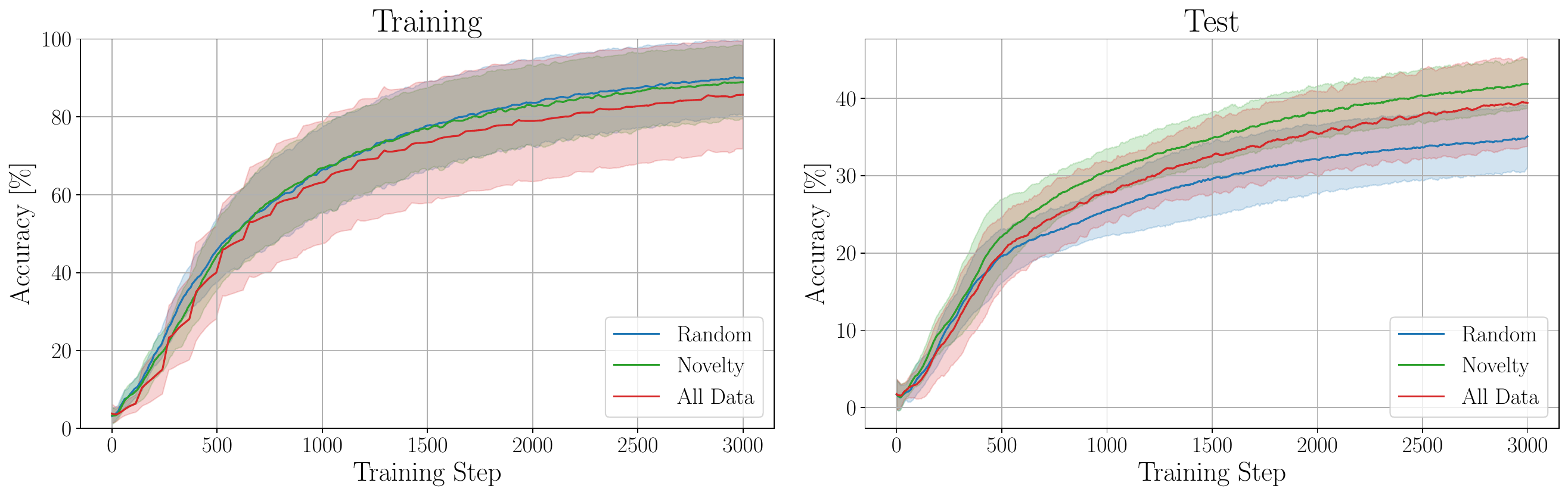}
    \caption{Training comparison of the classifier using the novelty filtered data (\mytikzLineGreen) with 82\% data reduction, random (\mytikzLineBlue) sampled data of the same size and all data (\mytikzLineRed) with RF 4. }
    \label{fig:training_process}
\end{figure*}
\begin{figure*}[t]
    \centering
    \includegraphics[width=0.94\linewidth]{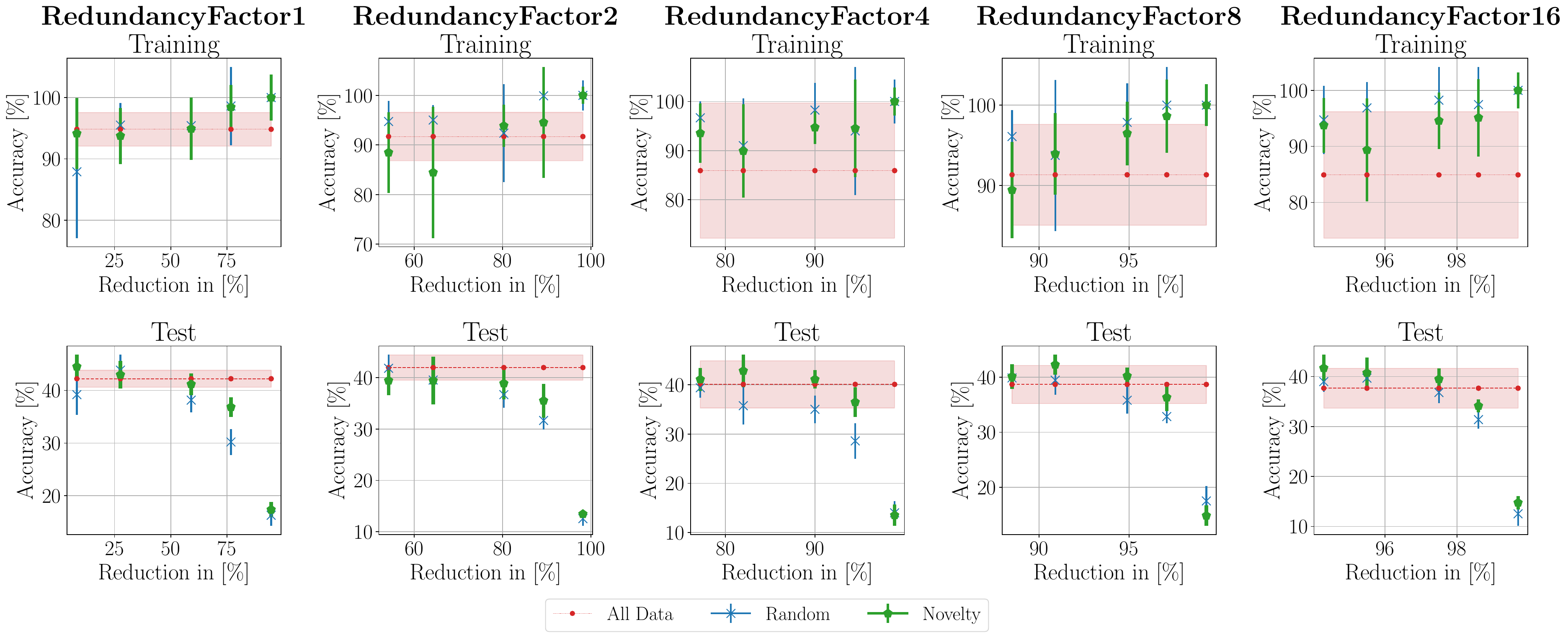}
    \caption{Training evaluation of a traffic sign classifier using different datasets including the novelty filtered data (\mytikzLineGreen)  of Mean Shift Algorithm, random (\mytikzLineBlue) sampled data of the same size and all data (\mytikzLineRed). The top Row shows the training accuracy, and the bottom row shows the test accuracy over the data reduction rate.}
    \label{fig:Evaluation}
\end{figure*}
To evaluate the impact of novelty filtering under varying redundancy and threshold conditions, a controlled data stream experiment was conducted using multiple replicated and thresholded subsets. Each image in the dataset belongs to a single class and is processed in a randomized stream.
For each incoming frame, the novelty score is computed using the dynamic Mean Shift algorithm. Based on a predefined threshold, images with low scores are discarded, while those with higher scores are retained and used to update the covariance model. This process is repeated across five Redundancy Factors (RF), 1×, 2×, 4×, 8×, and 16×, created by duplicating the dataset accordingly.
For each redundancy set, five different novelty score thresholds $\mathbf{T}$ (2500, 5000, 10000, 15000, 30000) are applied, resulting in 25 novelty-filtered datasets as a combination of redundancy dataset and novelty thresholds. For benchmarking, for each corresponding novelty-filtered dataset, a random dataset of the same size is retrieved from all data.  An additional dataset containing all images (no filtering) is also included for each redundancy level, totaling 55 datasets of all, random and novelty-filtered data across all redundancy levels.
The data selection process is illustrated in Fig.~\ref{fig:selection_process}. Initially, no samples are discarded within the first 78 frames. As the stream progresses, the filtering mechanism increasingly rejects incoming images, reflecting growing redundancy. By frame index 8743, more samples are discarded than retained due to their similarity to previously selected data. From frame index 52984 onward, all incoming frames are considered redundant. This behavior results from the system’s dynamic adaptation to evolving definitions of novelty.

\subsubsection{Training Evaluation}
\begin{figure*}[h]
    \centering
    \includegraphics[width=0.92\linewidth]{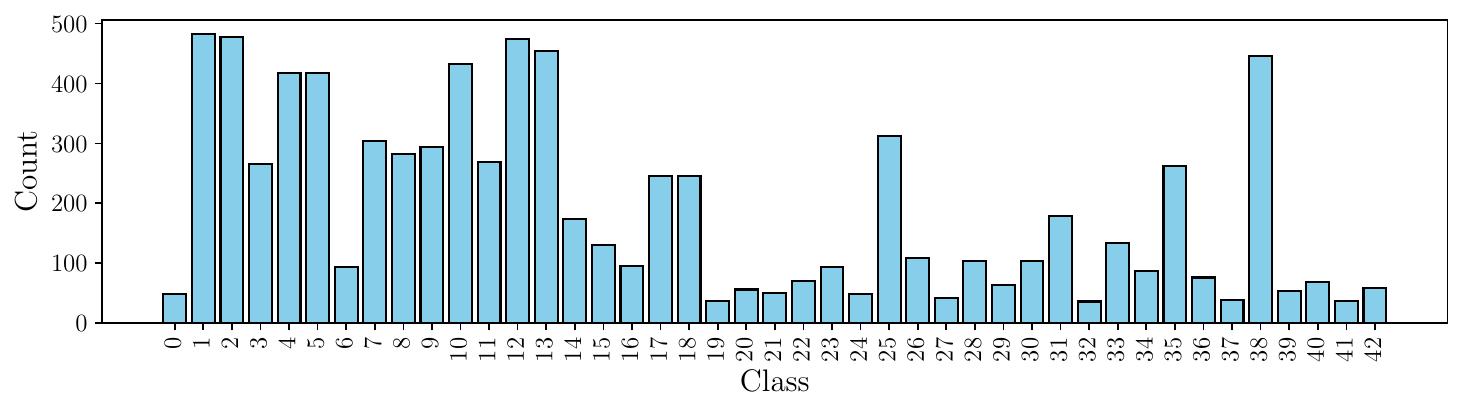}
    \caption{Distribution of the classes counts for each class category in the Traffic Sign dataset.}
    \label{fig:class_distribution}
\end{figure*}
\begin{table*}[t]
    \centering

\begin{adjustbox}{max width=\textwidth}
\begin{tabular}{l*{30}{c}}
\toprule
\toprule
\multirow{2}{*}{} 
& \multicolumn{5}{c}{\textbf{Reduction Rate (\%)}} \\
 & 78.09 & 84.00 & 92.15 & 96.34 & 99.48 & All \\
\midrule
CoV  $\downarrow$   
                     & 0.77 $\pm$ 0.00 &  0.73 $\pm$ 0.0   &   0.65 $\pm$ 0.01  &  0.58 $\pm$ 0.02   &  \textbf{0.42 $\pm$ 0.02}  & 0.79 $\pm$0.0 \\
Entropy $\uparrow$  
                     & 0.92 $\pm$ 0.00 &  0.93 $\pm$ 0.0   &   0.94 $\pm$ 0.0   &   0.95 $\pm$ 0.0  &  \textbf{ 0.98 $\pm$ 0.0} & 0.92 $\pm$0.0  \\
IR    $\downarrow$  
                     & 12.54  $\pm$ 0.34 &  12.65 $\pm$ 0.58   &   11.26 $\pm$ 0.96   &   9.67 $\pm$ 0.81  & \textbf{5.90 $\pm$ 1.24}  & 13.29 $\pm$0.0 \\
\bottomrule
\end{tabular}
\end{adjustbox}

    \caption{Evaluation of dataset balance derived by the novelty filter compared against all data and random sampling.}
    \label{tab:balance_table}
\end{table*}
\begin{table}[t]
    \centering
    
\resizebox{\columnwidth}{!}{%
\begin{tabular}{lcccccc}
\toprule\toprule
\multirow{2}{*}{Metric} & \multicolumn{3}{c}{RF4} \\
\cmidrule(lr){2-4} \cmidrule(lr){5-7}
 &Reduction (\%)  & Novelty & Random   \\
\midrule
\multirow{5}{*}{\shortstack{Avg. pairwise\\ Cosine Similarity} $\uparrow$} 
& 0     &   -           & $0.19\pm0.02$  \\
& 78.09 & $0.20\pm0.02$  &  $0.19\pm0.02$ \\
& 84.00 & $0.20\pm0.02$  & $0.19\pm0.02$ \\
& 92.15 & $0.21\pm0.02$  & $0.19\pm0.02$ \\
& 96.34 & $0.22\pm0.02$ &   $0.18\pm0.02$ \\
& 99.47 & $0.25\pm0.03$ &  $0.19\pm0.04$  \\
\midrule
    \multirow{5}{*}{\shortstack{Avg. pairwise \\ Euclidean Distance } $\uparrow$ } 
& 0 & -            & $454\pm31.9$ \\
& 78.09 & $460\pm 29.6$ & $454\pm 32.2$ \\
& 84.00 & $469\pm28.7$ & $454.\pm 28.9$ \\
& 92.15 & $482\pm27.9$ & $450\pm 35.3$ \\
& 96.34 & $493\pm30.5$ & $452\pm 30.1$ \\
& 99.47 & $524\pm30.5$ & $ 456\pm 53.1$ \\
\bottomrule
\end{tabular}
}

    \caption{Evaluation of dataset diversity derived by the novelty filter compared against all data and random sampling.}
    \label{tab:diversity}
\end{table}
To assess the impact of data reduction on neural network classifier performance, a traffic sign object detector was trained on each of the datasets generated in Sec. \ref{sec:dataReduction}. The training is conducted with five different seeds for robustness, resulting in a total of 255 training runs. The classifier architecture consists of two convolutional layers (12 and 18 channels), followed by two fully connected layers (12 and 32 neurons, respectively). 
The training is stopped after 6000 steps with a batch size of 128 using a learning rate of $1\mathrm{e}{-4}$ with Adam optimization. An exemplar training curve is shown in Fig. \ref{fig:training_process}, while the performance evaluation is presented in Fig. \ref{fig:Evaluation}. For each RF, the top row shows training accuracy, and the bottom row tests accuracy, both plotted against the level of data reduction.
The results show that dataset reduction via novelty filtering can improve generalization. This is especially clear in the test accuracy: in some cases, novelty-filtered datasets outperform the dataset containing all data, as indicated by green markers above the red baseline in Fig. \ref{fig:Evaluation} bottom. That is for instance, with RF 1, the test accuracy improves from 42\% to 44\% when 10\% of the data is discarded. With higher redundancy (e.g., factor 8), the model maintains or improves performance even when over 90\% of the data is discarded, highlighting the effectiveness of the novelty filtering.
Random sampling offers occasional performance gains but introduces higher variance and less consistent outcomes. While some settings show comparable test accuracy (e.g., RF 1 with 30\% reduction), the average performance across seeds is lower than novelty-based filtering. Additionally, random sampling often leads to higher training accuracy but lower test accuracy, indicating overfitting. If the data filtering is too aggressive, this leads to overfitting and a collapse in model performance.
Across all redundancy levels, the optimal training set size, yielding the best test performance, was consistently ~80\% of the original dataset size, achieved through novelty-based filtering. This demonstrates that the dynamic Mean Shift filtering reduces not only redundant data but also is able to enhance model performance.

\subsection{Diversity and Class Balance Evaluation}
The GTSB dataset used in this study is inherently imbalanced, as shown by the class distribution in Fig.~\ref{fig:class_distribution}. To evaluate how the Mean Shift-based novelty filtering affects class balance, three metrics for each reduced dataset is computed:
Coefficient of Variation (CV) measures the relative dispersion of class frequencies. Let $f_i$ denote the number of samples in class $i$, and $N$ the number of classes. Then:
\begin{equation*}
\mu = \frac{1}{N} \sum_{i=1}^{N} f_i, \quad 
\sigma = \sqrt{\frac{1}{N} \sum_{i=1}^{N} (f_i - \mu)^2}, \quad 
\text{CV} = \frac{\sigma}{\mu}
\end{equation*}
Normalized Entropy (NE) quantifies the uniformity of the class distribution. For class probabilities $p_i = f_i / \sum_{j=1}^{N} f_j$, the entropy $H$ is:
\begin{equation*}
H = -\sum_{i=1}^{N} p_i \log p_i, \quad 
\text{NE} = \frac{H}{\log N}
\end{equation*}
Imbalance Ratio (IR) captures the ratio between the largest and smallest class sizes:
\begin{equation*}
\text{IR} = \frac{\max_i f_i}{\min_i f_i}
\end{equation*}
For diversity evaluation, the pairwise average angle and distance in each class using cosine similarity ($\cos$) and Euclidean distance ($d$) is calculated:
\begin{equation*}
    \cos(\vec{A}, \vec{B}) = \frac{\vec{A} \cdot \vec{B}}{\|\vec{A}\| \|\vec{B}\|}
\end{equation*}
\begin{equation*}
    d(\vec{A}, \vec{B}) = \sqrt{\sum_{i=1}^{n} (A_i - B_i)^2}
\end{equation*}

The evaluation of the balance of the dataset reveals that with more aggressive filtering strategies, the datasets get more balanced, see table \ref{tab:balance_table}. The same behavior is true for the dataset diversity with more diverse datasets. More aggressive filtering leads to more diverse datasets, see table \ref{tab:diversity}. The improved class balance and diversity likely contribute to the observed performance gains in the classifier.  Although dataset balance and diversity improve with aggressive dataset filtering using the Mean Shift novelty detection, the strategy has to be chosen carefully since filtering too much can lead to a collapse of model performance, which was evaluated in the previous section.
\section{Conclusion \& Outlook}
\label{sec:conclusion_and_outlook}

A dynamic algorithm based on the mean shift algorithm is introduced for online novelty detection in video data streams. This data-driven algorithm approach identifies normal and novel objects within image frames. Results indicate that reducing the size of training data improves performance, whereas increased data redundancy tends to degrade it. However, higher redundancy allows for more aggressive and effective filtering.  Random sampling provides only limited benefits and often leads to overfitting and instability. Overall, it can be concluded that this method is suitable for  
efficient data selection, reducing data size while improving dataset quality. With 32 frames per second, the method supports real-time execution for online selection.
Future work will involve extending the method to scene-level novelty detection. This shift is expected to introduce distributional mismatch,  for which the use of mixture models in combination with the dynamic mean shift is currently being investigated.

\section{ACKNOWLEDGMENT}
This work results from the just better DATA (jbDATA)
project supported by the German Federal Ministry for Economic Affairs and Climate Action of Germany (BMWK) and the European Union, grant number 19A23003H.
\appendix 

\subsection*{A.1 Proof of the Adaptive Mean }\label{sec:appendix_A1}

The feature vector of $\mathrm{N}$ different frames of the normal dataset $\mathcal{N}$ is defined as follows
\begin{equation*}
    \begin{split}
    \bm{\mu}_\mathrm{N} = \frac{1}{N} \sum_{j=1}^{N}  \bm{z}_{j}.
    \end{split}
\end{equation*}
The adaptive update for the definition of the mean of an updated normal dataset $\mathcal{N'}$ can be computed as stated in \eqref{eq:cont_mu_short}.

\begin{proof}
If the mean feature vector is extended by $\mathrm{K}$ frames, with $\mathrm{M}=\mathrm{N}+\mathrm{K}$ this results in
\begin{equation*} \label{eq1}
\begin{split}
\bm{\mu}_{\mathrm{M}} &= \frac{1}{\mathrm{M}} \sum_{j=1}^{\mathrm{M}} \bm{z}_{j,i} \\
 & = \frac{1}{\mathrm{M}} \Bigg[  \sum_{j=1}^{N}  \bm{z}_{j,i} + \sum_{j=N+1}^{\mathrm{M}}  \bm{z}_{j,i} \Bigg]
\end{split}
\end{equation*}
Expanding the first sum with $\sfrac{N}{N}$ and the second sum with $\sfrac{K}{K}$ results in
\begin{equation*}
    \begin{split}
    \bm{\mu}_\mathrm{M} = \frac{1}{\mathrm{M}}& \Bigg[ \frac{N}{\mathrm{N}} \sum_{j=1}^{N}  \bm{z}_{j} +\frac{K}{K} \sum_{j=N+1}^{K}  \bm{z}_{j} \Bigg] 
    \end{split}
\end{equation*}
Defining the new feature vector for $K$ frames as
\begin{equation*}
    \bm{\bm{\mu}}_\mathrm{K} = \frac{1}{K} \sum_{j=N+1}^{K} \bm{z}_{j,i}
\end{equation*}
The definition of the mean feature vector of the normal data set $\bm{\mu}_\mathrm{N}$ results in the new mean feature vector of the normal data set
\begin{equation*}\label{mu_k}
    \bm{\bm{\mu}}_\mathrm{M} = \frac{1}{\mathrm{M}} \left[ N\bm{\mu}_\mathrm{N} + K\bm{\mu}_\mathrm{K} \right]
\end{equation*}
which allows the mean feature vector of the entire data set to be continuously recalculated.
\end{proof}

\subsection*{Proof of the Adaptive Covariance}\label{sec:appendix_A2}
The covariance of N different frames is defined  as follows
\begin{equation*}
    \bm{\hat{\Sigma}} = \frac{1}{\mathrm{N}-1} (\bm{z} - \bm{\mu}_{\mathrm{N}})^\intercal (\bm{z} - \bm{\mu}_{\mathrm{N}}).
\end{equation*}
The adaptive update for the definition of the covariance of the updated normal dataset $\mathcal{N'}$  can be computed as stated in \eqref{eq:cont_cov}.
\begin{proof}
Given $\mathrm{N}$ frames and the corresponding mean vector $\bm{\mu}_\mathrm{\mathrm{n}}\in \mathbb{R}^D$ of the normal dataset , the definition of the covariance is
\begin{equation*}
    \bm{\hat{\Sigma}} = \frac{1}{\mathrm{N}-1} \sum_{j=1}^{\mathrm{N}} (\bm{z}_j - \bm{\mu}_\mathrm{N})(\bm{z}_j - \bm{\mu}_\mathrm{N})^\intercal,
\end{equation*}
with $ \bm{\hat{\Sigma}} \in \mathbb{R}^{\mathrm{D}\times \mathrm{D}}$.
For updating $\bm{\hat{\Sigma}}$ with $\mathrm{K}$ frames for the updated dataset $\mathcal{N'}$, with  $\mathrm{M}=\mathrm{N}+\mathrm{K}$ results in
\begin{equation*}
    \bm{\hat{\Sigma}}_{\mathrm{M}} = \frac{1}{\mathrm{M}-1} \sum_{j=1}^{\mathrm{M}} (\bm{z}_j - \bm{\mu}_{\mathrm{M}})(\bm{z}_j - \bm{\mu}_{\mathrm{M}})^\intercal
\end{equation*}
following
\begin{equation*}
\begin{split}
    \bm{\hat{\Sigma}}_{\mathrm{M}} &= \frac{1}{\mathrm{M}-1} \sum_{j=1}^{\mathrm{M}} \left[ \bm{z}_j \bm{z}_j^\intercal - \bm{z}_j \bm{\mu}_{\mathrm{M}}^\intercal - \bm{\mu}_{\mathrm{M}} \bm{z}_j^\intercal + \bm{\mu}_{\mathrm{M}} \bm{\mu}_{\mathrm{M}}^\intercal \right]\\
    &= \frac{1}{\mathrm{M}-1} \Bigg[ \left( \sum_{j=1}^{\mathrm{M}} \bm{z}_j \bm{z}_j^\intercal \right) - \bm{\mu}_{\mathrm{M}} \left( \sum_{j=1}^{\mathrm{M}} \bm{z}_j \right)^\intercal \\
    &\qquad\qquad\qquad\qquad\quad- \left( \sum_{j=1}^{\mathrm{M}} \bm{z}_j \right) \bm{\mu}_{\mathrm{M}}^\intercal + \mathrm{M} \bm{\mu}_{\mathrm{M}} \bm{\mu}_{\mathrm{M}}^\intercal \Bigg]\\
    &= \frac{1}{\mathrm{M}-1} \Bigg[ \left( \sum_{j=1}^{\mathrm{N}} \bm{z}_j \bm{z}_j^\intercal + \sum_{j=\mathrm{N}+1}^{\mathrm{M}} \bm{z}_j \bm{z}_j^\intercal \right) \\
    &\qquad\qquad\qquad -\bm{\mu}_{\mathrm{M}} \left( \sum_{j=1}^{\mathrm{N}} \bm{z}_j + \sum_{j=\mathrm{N}+1}^{\mathrm{M}} \bm{z}_j \right)^\intercal \\
    &\qquad  -\left( \sum_{j=1}^{\mathrm{N}} \bm{z}_j + \sum_{j=\mathrm{N}+1}^{\mathrm{M}} \bm{z}_j \right) \bm{\mu}_{\mathrm{M}}^\intercal 
    + \mathrm{M} \bm{\mu}_{\mathrm{M}} \bm{\mu}_{\mathrm{M}}^\intercal \Bigg] 
\end{split}
\end{equation*}
Defining the sums
\begin{subequations}
    \begin{equation*}
        \bm{A}_{NS} = \sum_{j=1}^{\mathrm{N}} \bm{z}_j \bm{z}_j^\intercal,
    \end{equation*}
    \begin{equation*}
        \bm{b}_{NS} = \sum_{j=1}^{\mathrm{N}} \bm{z}_j
    \end{equation*}
\end{subequations}
this results in the covariance matrix $\bm{\hat{\Sigma}}_\mathrm{M}$ updated by K frames to
\begin{multline*}
    \bm{\hat{\Sigma}}_\mathrm{M} = \frac{1}{\mathrm{M}-1} \bigg[\bm{A_\mathrm{M}}
    - \bm{\mu}_\mathrm{M}\bm{b_\mathrm{M}}^\intercal \\
    - \bm{b}_\mathrm{M}\bm{\mu}_\mathrm{M}^\intercal 
    + \mathrm{M}\cdot\bm{\mu}_\mathrm{M}\bm{\mu}_\mathrm{M}^\intercal \bigg].
\end{multline*}
with 
\begin{subequations}
\begin{equation*}
    \bm{A}_{\mathrm{M}} = A_{N} + \sum_{j=\mathrm{N}+1}^{\mathrm{M}} \bm{z}_j \bm{z}_j^\intercal 
\end{equation*}
\begin{equation*}
    \bm{b}_{\mathrm{M}} = b_{N} + \sum_{j=\mathrm{N}+1}^{\mathrm{M}} \bm{z}_j
\end{equation*}
\end{subequations}
which allows an adaptive update of the covariance.
\end{proof}




\bibliographystyle{IEEEtran}
\bibliography{root}

\end{document}